\documentclass[fleqn,twoside,10pt]{article}
\usepackage{epstopdf}
\usepackage{epsfig}
\usepackage{natbib}
\usepackage{left_eq}   
\usepackage{escape26}  
\usepackage{dsfont}

\usepackage[table,xcdraw]{xcolor}
\usepackage{pstricks, pst-plot}	
\usepackage{graphicx} 
\usepackage{wrapfig}  
\usepackage[figuresright]{rotating}
\usepackage{subfigure}

\usepackage{fancyhdr,graphicx,amsmath,amssymb}
\usepackage[ruled,vlined]{algorithm2e}
\include{pythonlisting}

\usepackage{amsmath}
\usepackage{amssymb}

\usepackage[english]{babel}
\usepackage{blindtext}

\newcommand{\BigO}[1]{\ensuremath{\operatorname{O}\bigl(#1\bigr)}}

\title{Bayesian Optimization with Dimension Scheduling: Application to Biological Systems}
 \author[a]{Doniyor Ulmasov}
 \author[b]{Caroline Baroukh}
 \author[c]{Benoit Chachuat}
 \author[a,*]{Marc Peter Deisenroth}
 \author[a,*]{Ruth Misener}
 \affil[a]{Department of Computing, Imperial College London, UK}
 \affil[b]{INRA, Laboratory of Environmental Biotechnology, France}
 \affil[c]{Department of Chemical Engineering, Imperial College London, UK}
 \email{*m.deisenroth, r.misener@imperial.ac.uk; \texttt{+44 (0) 20759 48315}}
 
 \HeaderTitle{Bayesian Optimization with Dimension Scheduling}
 \HeaderAuthor{Ulmasov, Baroukh, Chachuat, Deisenroth, Misener}

\begin{document}

\maketitle             
\thispagestyle{empty}  


\begin{abstract}
Bayesian Optimization (BO) is a data-efficient method for global black-box optimization of an expensive-to-evaluate fitness function. 
BO typically assumes that computation cost of BO is cheap, but experiments are time consuming or costly. In practice, this allows us to optimize ten or fewer critical parameters in up to 1,000 experiments. But experiments may be less  expensive than BO methods assume: In some simulation models, we may be able to conduct multiple thousands of experiments in a few hours, and the computational burden of BO is no longer negligible compared to experimentation time. To address this challenge we introduce a new Dimension Scheduling Algorithm (DSA), which reduces the computational burden of BO for many experiments. The key idea is that DSA optimizes the fitness function only along a small set of dimensions at each iteration. This DSA strategy (1) reduces the necessary computation time, (2) finds good solutions faster than the traditional BO method, and (3) can be parallelized straightforwardly. We evaluate the DSA in the context of optimizing parameters of dynamic models of microalgae metabolism and show faster convergence than traditional BO.


\end{abstract}


\section{Introduction}
Bayesian Optimization (BO) is a data-efficient method for global black-box optimization of an expensive-to-evaluate fitness function. The working hypothesis of BO is that experiments are very expensive, e.g., in terms of time or money, while BO computations are relatively cheap. 
BO has been applied to a wide variety of problems, including tuning critical parameters of Deep Neural Networks~\citep{Dahl2013}, learning controller parameters of walking robots~\citep{Calandra2015a} or automatic algorithm configuration~\citep{Hutter2011,Snoek2012}. 
To balance exploration and exploitation during optimization BO uses Gaussian Processes (GPs) to model a posterior distribution over fitness functions from available experiments~\citep{Jones:1998}. Similar to experimental design, an acquisition function is applied to the GP posterior over fitness functions to suggest the next (optimal) experiment. In all applications, there is a common theme: BO performance degrades in high dimensions and/or with large data amount of observation points due to the scaling of the Gaussian Processes (GP), the surrogate functions. In practice, BO is limited to optimizing about ten parameters and about 1,000 experiments since  GP predictions/evaluations scale linearly in the number of dimensions but cubically in the number data points when we integrate out the hyperparameters. 

Dynamic models of biological processes allow us to test biological hypotheses while running fewer costly, real-world experiments \citep{floudas-pardalos:2000}. This paper considers estimating biological parameters (e.g., reaction rate kinetics) by minimizing the squared error between model and experimental data points \citep{rodrigues:2006b}. We propose BO for efficient parameter estimation of a dynamic microalgae metabolism model \citep{baroukh-etal:2014}. The forcing function is based on light exposure and nitrate input; experimental data has been collected for measurable outputs including lipids, carbohydrates, carbon organic biomass, nitrogen organic biomass and chlorophyll. But our method is general and may be applied to any process model. 

There are several timescales for collecting microalgae metabolism data: an experiment of \cite{lacour-etal:2012} may take 10 days while each model simulation of \cite{baroukh-etal:2014} runs in a fraction of a second. BO is traditionally applied to functions with an expensive evaluation costs, e.g., running a 10 day experiment, but the objective of this paper is testing biological hypotheses; we are specifically interested in running the simulation model many times for parameter estimation. 

Parameter estimation for dynamic systems has been alternatively proposed using deterministic global optimization \citep{michalik-etal:2009}, heuristics \citep{rodrigues:2006a}, and using gradient-based methods. Problems of this size are, practically speaking, out of the range of current deterministic global optimization technology, so we are left to consider heuristics and gradient-based approaches. We prefer Bayesian optimization to heuristics such as genetic algorithms because the GPs allow us to immediately study sensitivity in the model response surface. Gradient-based methods directly benefit from the proposed Bayesian optimization approach since we can warm start the gradient optimization at promising initialization points.

Evaluating the microalgae metabolism model in our application so quickly leads to data set sized standard BO cannot manage. To address this problem we introduce Bayesian Optimization with Dimension Scheduling Algorithm (DSA). The DSA distributes the training data across many GPs, with each GP containing training data of a subset of the dimensions. At each iteration, a new subset of the dimensions is sampled from a probability distribution, which reflects the importance of the corresponding parameters, and the utility or acquisition function is optimized with respect to this subset. DSA benefits from a faster computational performance because each iteration considers a relatively small number of prior experiments. 


\section{Background and Related Work}\label{sec:background}

Bayesian Optimization (BO)~\citep{Kushner1964} is a global black-box technique for optimizing $f: \mathbb{R}^d \mapsto \mathbb{R}$, which may be non-convex, on a $d$-dimensional hyper-rectangle with finite bounds $B$~\citep{Jones:1998}:
\begin{align}
x^* \in\arg\min_{x \in B \subset \mathds{R}^{d}} f(x)\,.
\end{align}
%
In the BO context, we consider a regression problem $y=f(x)+\epsilon$, where $x\in\mathds{R}^d$, $y\in\mathds{R}$, and $f$ is the unknown fitness/utility function. The Gaussian likelihood $p(y|f(X))=\mathcal{N}(f(x),\sigma_\epsilon^2)$ accounts for the independently and identically distributed measurement noise $\epsilon\sim\mathcal{N}(0,\sigma_\epsilon^2)$. The objective is to infer the latent utility function $f$ from a training data set $X = \{x_i\}_{i=1}^N, Y = \{y_i\}_{i=1}^N$.
%
A GP is defined as a collection of random variables, any finite number of which is Gaussian distributed. A GP is fully specified by a mean function $m$ and a covariance function $k$ (kernel) with hyper-parameters $\psi$, which allows us to encode high-level structural assumptions, e.g., differentiability. Without loss of generality, we assume that the prior mean function is 0. 
A GP is trained by finding hyper-parameters $\theta =
\{\psi, \sigma_\epsilon\}$ that maximize the log-marginal
likelihood
\begin{align}
  \log p(Y|X, \theta)=  -\tfrac{1}{2}\big(Y^T(K+\sigma_\epsilon^2 I)^{-1}Y + \log|K+\sigma_\varepsilon^2 I|\big) + \text{ const}\,,
\label{eq:log-marginal likelihood}
\end{align}
where $K=k(X,X)\in\mathds{R}^{N\times N}$ is the kernel
matrix.  

For a given set of hyper-parameters $\theta$, a training set $X, Y$ and a test input $x_*\in\mathds{R}^d$, the GP posterior predictive distribution of the corresponding function value $f_* = f(x_*)$ is Gaussian with mean and variance given by
\begin{align}
  \mathrm{E}[f_*] =  k_*^T(K + \sigma_\varepsilon^2 I)^{-1}Y\,,\qquad \mathrm{var}[f_*] = k_{**} - k_*^T(K + \sigma_\varepsilon^2 I)^{-1} k_*\,,
\end{align}
respectively, where we defined $k_* = k(X,x_*)$ and $k_{**}=k(x_*, x_*)$. 

%
BO places a GP prior on the fitness function $f$, which serves as a probabilistic surrogate model for $f$. Instead of optimizing $f$, which will require many function evaluations, BO optimizes an acquisition function $a(\cdot)$ that depends on the GP posterior predictive distribution to decide the next evaluation point (experiment).\footnote{In this paper, we choose Expected Improvement~\citep{Mockus1978} as acquisition function and use DIRECT~\citep{Jones:1993} to optimize it.} This experiment yields an additional training point $(x_i, y_i)$, which updates the GP model. After updating the GP with the new observation, BO repeats the cycle until convergence or an upper bound on the total number of experiments. BO is summarized in Alg.~\ref{alg:typical_bo}.
\begin{algorithm}[tb]
\label{alg:typical_bo}
\SetAlgoLined
Evaluate the objective function $f$ $N$ times; Obtain initial training set $(X, \, Y)$\;
\For{$n=N$ to \text{maxIter}}
{
  Update the GP with $(X, \, Y)$\;
  $x_{n+1} = \arg \min_x~a(x|GP_n)$\;
  Evaluate objective function $y_{n+1} = f(x_{n+1}) + \varepsilon$\;
  Augment training set $X,Y$ with $(x_{n+1}, \, y_{n+1})$\;
}
\caption{Bayesian Optimization}
\end{algorithm}
%
%
%
There have been approaches toward scaling BO to more dimensions. The Additive BO method (ABO) developed by \cite{kandasamy2015high} is the variant of BO most similar to our DSA contribution: ABO assumes that the objective function $f(x)$ can be decomposed into $M$ functions $f(x) = \sum_{k=1}^m f^k(x^k)$ operating on mutually disjoint dimensions, such that $x^i \cap x^j = \emptyset $ for $i \ne j$. Because of the decomposability assumption, each of the $M$ functions can be optimized separately. ABO expedites typical BO approaches by explicitly breaking dependence between the functional domains. Although \cite{kandasamy2015high} show that their method is applicable to non-additive functions we find that, in the case of our parameter estimation problem, we do not get good results using ABO. We now introduce DSA, which shares the ABO goal of cutting BO computation time but abandons the assumption of functional decomposability and, in particular, the idea of mutually disjoint parameter domains.

\section{Bayesian Optimization with Dimension Scheduling Algorithm}\label{sec:dsa}

\begin{algorithm}[tb]
\label{alg:dsa}
\SetAlgoLined
 Evaluate objective function $N$ times, update the all GPs with sampled data $(X, \, Y)$\; 
 Set $x_b =\arg\min f(X)$ and $y_b =\min Y$\;
 \For{$n=N$ to \text{maxIter}}{
  Update probability vector $P$ from the observations\;
  Randomly sample dimension set $Z$ from $P$\;
  Find $x_{n+1}^Z=\arg\min~a(x_{n+1}^Z, \, GP_Z)$\;
  Replace coordinates $Z$ of $x_{n+1}$ with $x_{n+1}^Z$\;
  Evaluate objective function $y_{n+1} = f(x_{n+1})+\varepsilon_{n+1}$\;
  Augment training set of $GP_Z$ with $(x_{n+1}^Z, \, y_{n+1}^Z)$\;
\textbf{if} $y_{n+1} < y_b$ \textbf{then} $(x_b, \, y_b) = (x_{n+1}, \, y_{n+1})$\;
}
\caption{Bayesian Optimization with Dimension Scheduling Algorithm}
\end{algorithm}
The DSA algorithm is an extension of the traditional BO algorithm and is summarized in Alg.~\ref{alg:dsa}.
The initialization corresponds to the standard procedure of BO and observed function values for a few sampled $x_i$. Based on the set of initial samples $X$ and corresponding observations $Y$  we set $x_b = \arg\min_X Y$ and $y_b =\min(Y)$, where $x_b$ stores our best known argument for the best objective value, and $y_b$ stores the best objective value. 
DSA samples coordinates from a probability vector $P$, which represents the relative importance of the $x$ coordinates. For this purpose, we apply PCA to the available data $X$ and set $P$ proportional to the eigenvalue magnitude every 50 iterations. DSA now samples a random dimension set $Z$ from  $P$. We now optimize the $Z$-coordinates of $x$ while clamping the other coordinates to their current best values. For  optimization we use a GP model $GP_Z$, which possesses as input dimensions only the $Z$-coordinates of $X$. If $GP_Z$ does not yet exists, we spawn a new GP model where we use the relevant coordinates of the initial samples $(X,Y)$ as training data. For the optimization, we optimize the acquisition function $a(\cdot)$ related to $GP_Z$ and obtain $x_{n+1}^{(Z)}$. We perform an experiment with the new $x_b$ and if the observed function value $y_{n+1}$ is a better objective value than $y_b$, we update $y_b$ to $y_{n+1}$ and $x_b$ to $x_{n+1}$. Note that we effectively only update the dimensions $Z$ in $x_b$ with parameter values from $x_{n+1}^Z$. The new data point $(x_{n+1}, \, y_{n+1})$ is only added to $GP_Z$, i.e., the data set size of the other GP models does not increase.

\section{Results}
\begin{table}[tb]
\centering
\caption{Best achieved objective in four experiments using standard BO and DSA. Each experiment ran for 500 iterations}
\label{tbl:experiments}
\begin{tabular}{
>{\columncolor[HTML]{FFFE65}}l 
>{\columncolor[HTML]{FFCC67}}l 
>{\columncolor[HTML]{FFCCC9}}l 
>{\columncolor[HTML]{9AFF99}}l 
>{\columncolor[HTML]{FFCCC9}}l 
>{\columncolor[HTML]{9AFF99}}l 
>{\columncolor[HTML]{FFCCC9}}l 
>{\columncolor[HTML]{9AFF99}}l 
>{\columncolor[HTML]{FFCCC9}}l 
>{\columncolor[HTML]{9AFF99}}l }
\multicolumn{1}{c}{\cellcolor[HTML]{FFC702}\textbf{Model}} & \multicolumn{1}{c}{\cellcolor[HTML]{FFC702}\textbf{d}} & \multicolumn{1}{c}{\cellcolor[HTML]{FFC702}\textbf{BO:1}} & \multicolumn{1}{c}{\cellcolor[HTML]{FFC702}\textbf{DSA:1}} & \multicolumn{1}{c}{\cellcolor[HTML]{FFC702}\textbf{BO:2}} & \multicolumn{1}{c}{\cellcolor[HTML]{FFC702}\textbf{DSA:2}} & \multicolumn{1}{c}{\cellcolor[HTML]{FFC702}\textbf{BO:3}} & \multicolumn{1}{c}{\cellcolor[HTML]{FFC702}\textbf{DSA:3}} & \multicolumn{1}{c}{\cellcolor[HTML]{FFC702}\textbf{BO:4}} & \multicolumn{1}{c}{\cellcolor[HTML]{FFC702}\textbf{DSA:4}} \\
\textbf{M19} & 10 & 54.62 & 32.32 & 52.66 & 47.20 & 39.35 & 30.46 & 41.70 & 24.48 \\
\textbf{M19o} & 10 & 77.78 & 66.09 & \cellcolor[HTML]{9AFF99}114.78 & \cellcolor[HTML]{FFCCC9}354.81 & \cellcolor[HTML]{9AFF99}78.64 & \cellcolor[HTML]{FFCCC9}133.36 & 80.05 & 69.54 \\
\textbf{M26} & 11 & 87.91 & 35.86 & 32.56 & 26.66 & 55.40 & 31.53 & 57.58 & 26.54 \\
\textbf{M29G} & 11 & \cellcolor[HTML]{9AFF99}46.32 & \cellcolor[HTML]{FFCCC9}59.77 & 72.19 & 35.16 & 86.61 & 77.40 & 73.62 & 64.59 \\
\textbf{M29C} & 11 & 36.50 & 24.78 & 44.71 & 37.25 & 41.47 & 34.59 & 52.74 & 33.77 \\
\textbf{M31} & 11 & 39.24 & 24.37 & 60.39 & 29.75 & 49.52 & 26.26 & \cellcolor[HTML]{9AFF99}44.18 & \cellcolor[HTML]{FFCCC9}48.53 \\
\textbf{M32C} & 12 & 43.33 & 33.14 & 52.19 & 31.15 & 50.62 & 28.02 & 51.13 & 26.18 \\
\textbf{M32G} & 12 & 42.51 & 27.12 & 47.55 & 26.78 & 52.38 & 31.45 & 53.21 & 23.54 \\
\textbf{M33} & 12 & 44.07 & 24.82 & 38.78 & 24.64 & 40.26 & 29.09 & 46.68 & 20.61 \\
\textbf{M35} & 12 & 50.16 & 25.86 & 57.72 & 22.86 & 57.06 & 28.97 & 57.00 & 40.11
\end{tabular}
\end{table}


We focus on parameter estimation for dynamic models of microalgae metabolism \citep{baroukh-etal:2014}. The  objective is to minimize a weighted sum of the squared errors between the model simulations and the real world experimental data. The weights were chosen as the mean of each kind of experimental data so that each kind of measurement (biomass, lipids, etc.) contributes equally to the error. Each model simulation is evaluated in a fraction of the second, but the wide bounds and the number of dimensions creates a large search area which prohibits random sampling methods.

\begin{figure}[tb]
\centering
\includegraphics[width=0.7\textwidth]{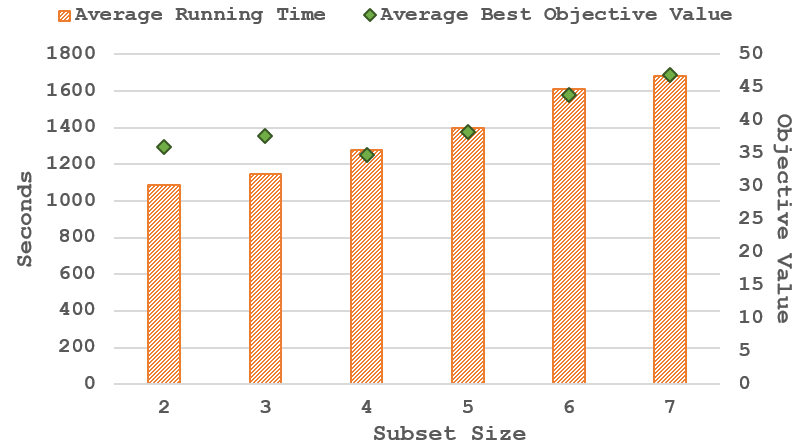} 
\label{subfig:subsets}
\caption{Average running time best objective value of different subset sizes with respect to five averaged runs of M19. Lower is better.}
\end{figure}

To compare traditional BO \citep{hoffman-shahriari:2014} and DSA, we tested 10 different  model variants with 4 experiment runs per model. Each model variant, e.g., M19, represents a different biological hypothesis on microalgae metabolism \citep{baroukh-etal:2014} and may have a varying number of parameters or possible parameter bounds. Each experiment ran for 500 iterations, with the same set of randomly sampled data, a Squared Exponential kernel and the Expected Improvement acquisition function. In all cases we chose dimension size 2 for the DSA. Figure~\ref{subfig:subsets} motivates using DSA dimension 2 by diagramming the average results from 5 experiments on model M19; note that the average running time and objective values are consistently better for dimension 2. 
Table~\ref{tbl:experiments} summarizes model details and results of the 4 runs, the column $d$ stands for dimensions in the model. 

\begin{figure}[tb]
\centering
\includegraphics[width=0.8\textwidth]
{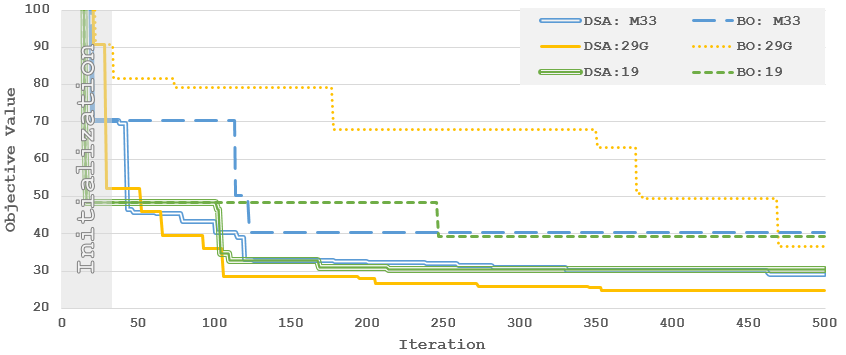}
\label{subfig:best_obj}
\caption{Best running objective value of DSA and classical BO for three models. Lower is better.}
\end{figure}

\begin{figure}[tb]
\centering
\includegraphics[width=0.9\textwidth]{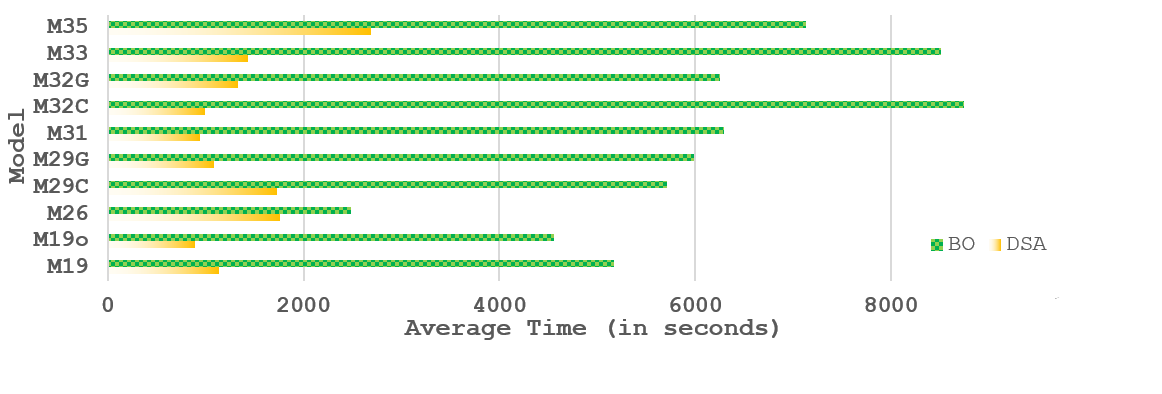} \label{subfig:ave_comput}
\caption{Average computation time of DSA and traditional BO for all models. DSA results in a substantial computational speed-up.}
\end{figure}

In most cases, DSA terminated with a lower objective value than traditional BO. The exceptions arise due to the limitations of DSA discussed in the next section. Figure \ref{subfig:best_obj} presents a sample of running best objective values. The traditional BO method tends to improve less frequently as the optimization process progresses. The DSA method, due to the constant changes of the dimensions, tends to improve the objective value more frequently and  achieves an overall lower objective value. Due to random sampling, the DSA performance is not uniformly superior to traditional BO, as seen in Table \ref{tbl:experiments}.

The DSA algorithm completes each experiment on average at a fifth of the computation time of BO: Figure~\ref{subfig:ave_comput} shows the average computation time of the experiment runs with all the models. The performance increase stems from the reduced number of GP dimensions and reduced number of observations per GP. The computational complexity of the predictions are still \BigO{n^2}. However,  by distributing the training data across multiple GPs the training sets per GP are relatively small (proportional to how often a set of dimensions $Z$ has been sampled from the probability vector $P$). The  number of GPs used by the algorithm is upper bounded by the total number of permutations of the set $Z$, which itself is based on the subset sizes and problem dimensionality. 

\section{Discussion}

Since the DSA algorithm never uses all of the data, we 
cannot assert global algorithm convergence. The issue of convergence could be addressed, for example, by a hybrid BO solution with the function space sampled for near optimal solutions with DSA.

The DSA addresses only a subset of problems with BO under specific conditions. The algorithm faces similar challenges when we increase number of total dimensions $d$, and the GP's accuracy would suffer since the observation points would be spread thinly over many GPs. In the current implementation, the maximum number of GPs is equivalent to number of permutations of the subset $Z$ used by the DSA. However, not all GPs are equally relevant, and some pruning may be beneficial to limit the number of models. The marginal likelihood could be used for this purpose. 

One of the advantages of the Bayesian Optimization with Dimension Scheduler over traditional Bayesian Optimization is a fairly straightforward code parallelization for increased performance on multicore systems. The traditional Bayesian Optimization method can be parallelized to a certain extent. There has been some work on Gaussian process models in distributed systems \citep{deisenroth-ng:2015, Gal2014}, which allows GPs to scale to larger data sets. The solvers can be parallelized by dividing the search space between different processes. These approaches provide compartmental parallelization, but the whole process is still sequential. 

The DSA parallelization process is much simpler, and can use current GP and solver modules. The solution lies in GP distribution across many processes, and a manager process to communicate between each child-process and the objective function. The manager would assign different iteration points to each process. Each process would contain a GP and a solver, and based on the iteration number the process receives (if we have exploration parameter scheduler), the process maximizes the acquisition function for the GP in the process and returns the solution. The manager would evaluate the solution and return it to the process to update the GP and start a new iteration. 

\section{Conclusion}
We have identified a performance issue with traditional BO when the experimentation time is significantly less than the optimization time and have therefore developed DSA. In the case of the microalgae model, DSA out-performed traditional BO in terms of computation time and the best objective value. The DSA could be applied to other models where a relatively quick computational model with high dimensionality requires efficient parameter estimation. Further DSA improvement, such as parallel implementation and GPs reduction based on the marginal likelihood would lead to even greater performance benefits; the parallelization potential suggests that the algorithm may be able to manage more dimensions than traditional BO. 




\noindent
\textbf{\large Acknowledgments}

This work was supported by the Engineering and Physical Sciences Research Council [EP/M028240/1] and a Royal Academy of Engineering Research Fellowship to R.M.

{\small\bibsep=0pt
\bibliography{BayesOpt_DSA}
}
\end{document}